\documentclass[letterpaper, 10 pt, conference]{ieeeconf}  
\pdfoutput=1

\IEEEoverridecommandlockouts                              

\usepackage{hyperref}
\usepackage{graphicx} 
\usepackage{amsmath} 
\usepackage{amssymb}  
\usepackage{bbm}
\usepackage{verbatim} 
\usepackage{algpseudocode}
\usepackage{tabulary}
\usepackage{algorithm} 
\usepackage[utf8]{inputenc} 
\usepackage{eso-pic}

\title{\LARGE \bf
Black-Box Data-efficient Policy Search for Robotics
}

\author{Konstantinos Chatzilygeroudis, Roberto Rama, Rituraj Kaushik, Dorian Goepp,\\Vassilis Vassiliades and Jean-Baptiste Mouret*\\
\thanks{\scriptsize*Corresponding author: {\tt\footnotesize jean-baptiste.mouret@inria.fr}}
\thanks{\scriptsize All authors have the following affiliations:}
\thanks{\scriptsize - Inria, Villers-lès-Nancy, F-54600, France}%
\thanks{\scriptsize - CNRS, Loria, UMR 7503, Vandœuvre-lès-Nancy, F-54500, France}%
\thanks{\scriptsize - Université de Lorraine, Loria, UMR 7503, Vandœuvre-lès-Nancy, F-54500, France}%
\thanks{\scriptsize This work received funding from the European Research Council (ERC) under the European Union's Horizon 2020 research and innovation programme (GA no. 637972, project ``ResiBots'') and the European Commission through the project H2020 AnDy (GA no. 731540).}%
}

\usepackage{color}
\DeclareMathOperator*{\argmax}{\arg\!\max}

\begin{document}
\AddToShipoutPicture*{\put(0,740){\parbox[b][\paperheight]{\paperwidth}{%
\vfill
\centering\footnotesize
Chatzilygeroudis, K., Rama, R., Kaushik, R., Goepp, D., Vassiliades, V., \& Mouret, J.-B. (2017)\\``Black-Box Data-efficient Policy Search for Robotics''\\Proceedings of the IEEE/RSJ International Conference on Intelligent Robots and Systems (IROS)
}}}
\maketitle
\thispagestyle{empty}
\pagestyle{empty}

\begin{abstract}
  The most data-efficient algorithms for reinforcement learning (RL) in robotics are based on uncertain dynamical models: after each episode, they first learn a dynamical model of the robot, then they use an optimization algorithm to find a policy that maximizes the expected return given the model and its uncertainties. It is often believed that this optimization can be tractable only if analytical, gradient-based algorithms are used; however, these algorithms require using specific families of reward functions and policies, which greatly limits the flexibility of the overall approach. In this paper, we introduce a novel model-based RL algorithm, called Black-DROPS (Black-box Data-efficient RObot Policy Search) that: (1) does not impose any constraint on the reward function or the policy (they are treated as \emph{black-boxes}), (2) is as data-efficient as the state-of-the-art algorithm for data-efficient RL in robotics, and (3) is as fast (or faster) than analytical approaches when several cores are available. The key idea is to replace the gradient-based optimization algorithm with a parallel, black-box algorithm that takes into account the model uncertainties. We demonstrate the performance of our new algorithm on two standard control benchmark problems (in simulation) and a low-cost robotic manipulator (with a real robot).
%
%
\end{abstract}


\section{Introduction}
Reinforcement Learning (RL) can help robots adapt to  unforeseen situations, such as being damaged~\cite{carlson_how_2005,cully2015robots,chatzilygeroudis2016resetfree} or stranded~\cite{nagatani2013emergency}. Rather than aborting their mission when something goes wrong, they could carry on by discovering new behaviors autonomously. Nevertheless, to be useful in such situations, learning has to happen in a few minutes, typically within a few trials.
This scarcity of data makes it difficult to exploit the many recent machine learning techniques (e.g., deep learning) that rely on the availability of very large datasets or fast simulations\footnote{As an illustration, the deep Q-learning algorithm needed about 38 days of interaction to learn to play Atari 2600 games~\cite{mnih_human-level_2015} (only four possible actions), which would be hardly conceivable to a achieve with a robot.}~\cite{lecun2015deep}. As a consequence, robot learning has to consider other approaches, with the explicit goal of requiring as little \emph{interaction time} as possible between the robot and the environment.
%
%

When data are scarce, a general principle is to extract as much information as possible from them. In the case of robotics, this means that all the state variables that are available should be collected at every time-step and be used by the learning algorithm. This contrasts with many direct policy search approaches (e.g., policy gradient algorithms~\cite{deisenroth_survey_2013,kohl2004policy} or Bayesian optimization~\cite{cully2015robots,calandra2015bayesian,lizotte2007automatic}) which only use the (cumulative) reward at the end of each episode.

One of the best ways to take advantage of this sequential state recording is to learn a dynamical model of the robot~\cite{nguyen2011model}, 
and then exploit it either for model-predictive control~\cite{camacho2013model} or to find an optimal policy offline~\cite{deisenroth_survey_2013}. However, such approaches assume that the model is ``good enough'' to predict future states for all the possible states. This is often not the case when only a few episodes have been performed, as many states have not been observed yet. Learning with a dynamical model therefore often requires acquiring enough points to learn an accurate model, which, in turn, increases the interaction time.

This challenge can be overcome by taking into account the uncertainty of the dynamical model: if the algorithm ``knows'' that a prediction is unreliable, it can balance the risks of trying something that might fail with the potential benefits. The PILCO (Probabilistic Inference for Learning COntrol) algorithm~\cite{deisenroth_gaussian_2015}, which is one of the state-of-the-art algorithms for data-efficient model-based policy search, follows this strategy by alternating between two steps, (1) learning a dynamical model with Gaussian processes~\cite{rasmussen2006gaussian}, (2) using a gradient-based optimizer to search for a policy that maximizes the expected reward, taking the uncertainty of the model into account.
Thanks to this process, PILCO achieves remarkable data-efficiency.

Nevertheless, analytical algorithms like PILCO have two main issues that may not be apparent at first sight. 
First, they impose several constraints on the reward functions and policies that prevent the use of arbitrary rewards (e.g., PILCO can only be used with distance-based rewards so far) and of non-derivable policies (e.g., parameterized state automata, like in~\cite{calandra2015bayesian}). 
Second, they require a large \emph{computation time} to optimize the policy (e.g., typically more than 5 minutes on a modern computer between each episode for the cart-pole benchmark), because they rely on computationally expensive methods to do approximate inference 
for each step of the policy evaluation~\cite{deisenroth_gaussian_2015}.

In this paper, we introduce a novel policy search algorithm that tackles these two problems while maintaining the data-efficiency of analytical algorithms. Our main insight is that while the analytic approach is efficient on a sequential computer, it cannot take advantage of the multi-core architectures now present in every computer. By contrast, Monte Carlo approaches and population-based black-box optimizers like CMA-ES~\cite{hansen2001completely} (1) do not put any constraint on the reward functions and policies, and (2) are straightforward to parallelize, which can make them competitive with analytical approaches when several cores are available. Our second insight is that it is not necessary to explicitly compute accurate approximations of the expected reward when the optimization is performed with rank-based algorithms designed for noisy functions (e.g., CMA-ES \cite{hansen2001completely}), which saves a lot of computation: only the ranking of potential solutions matters. Thus, it is possible to define a data-efficient, black-box policy search algorithm that is competitive with gradient-based, analytical approaches.
%
%

We call our algorithm Black-DROPS, for Black-box Data-efficient RObot Policy Search. It is a model-based policy search algorithm which:
\begin{itemize}
  \item takes into account the uncertainty of the dynamical model when searching for a policy;
  \item is as data-efficient as state-of-the-art, analytical algorithms, that is, it requires similar \emph{interaction time};
  \item performs a more global search than gradient-based algorithms, that is, it can escape from some local optima;
  \item is at least as fast as state-of-the-art, analytical methods when several cores are used, that is, it requires similar or lower \emph{computation time}; in addition, it is likely to be faster with future computers with more cores;
  \item does not impose any constraint on the reward function (in particular, the reward function can be learned);
  \item does not impose any constraint on the policy representation (any parameterized policy can be used).
\end{itemize}
We demonstrate these features with two families of policies, feed-forward neural networks and Gaussian processes, applied to two classic control benchmarks in simulation, the inverted pendulum and the cart-pole swing-up, as well as a physical 4-DOF robotic arm.

\section{Related Work}

Direct policy search (PS) methods have been successful in robotics as they can easily be applied in high-dimensional continuous state-action RL problems~\cite{deisenroth_survey_2013}. REINFORCE~\cite{williams1992simple} is an early policy gradient method which performs exploration of the action space using probabilistic policies. It suffers, however, from slow convergence due to the high variance in its gradient estimates. Policy gradients with parameter-based exploration (PGPE)~\cite{sehnke2008policy} address this problem by transferring exploration to parameter space. In particular, PGPE samples deterministic policies at the start of each episode by maintaining a separate Gaussian distribution for each parameter of the policy, whose mean and variance are adapted during training. The PoWER (Policy learning by Weighting Exploration with the Returns) algorithm~\cite{kober2011power} uses probability-weighted averaging, which has the property of following the natural gradient without computing it~\cite{kober2011power}. PoWER, however, assumes that the immediate rewards sum to a constant number and are always positive, which complicates the design of reward functions. The Policy Improvements with Path Integrals (PI$^2$)~\cite{theodorou2010generalized} algorithm does not make such an assumption. When the reward function is compatible with both PoWER and PI$^2$, the algorithms have identical performance~\cite{theodorou2010generalized}.

A limitation of PGPE is that it does not consider any correlations between dimensions in parameter space. This can be addressed by the Natural Evolution Strategies (NES)~\cite{wierstra2014natural} and Covariance Matrix Adaptation ES (CMA-ES)~\cite{hansen2001completely} families of algorithms, which are population-based Black-Box Optimizers (BBO). Both NES and CMA-ES iteratively update a search distribution by calculating an estimated gradient on the distribution parameters (mean and covariance matrix). At each generation, they sample a set of solutions (i.e., policy parameters) and rank them based on their fitness (i.e., expected return). NES performs gradient ascent along the natural gradient, which normalizes the update with respect to uncertainty. CMA-ES updates the distribution by exploiting the technique of evolution paths to average-out random effects over the generations. NES and CMA-ES are closely related, as the latter performs an approximate natural gradient ascent~\cite{akimoto2010bidirectional}. Interestingly, a variant of PI$^2$ with a simplified parameter perturbation and update method outperforms PI$^2$ and was shown to be a special case of CMA-ES~\cite{stulp2013robot}.

In general, any BBO can be used for direct PS. Bayesian optimization~\cite{shahriari_taking_2016} is a particular family of BBO that can be very data-efficient by building a surrogate model of the objective function (i.e., the expected return) and exploring this model in a clever way (e.g., using upper confidence bounds~\cite{auer2002using}). It can drastically decrease the evaluation time when optimizing gaits~\cite{lizotte2007automatic} or when finding compensatory behaviors for damaged robots~\cite{cully2015robots}.

The data-efficiency of direct PS can be further increased by learning the model (i.e., transition and reward function) of the system from data and inferring the optimal policy from the model~\cite{deisenroth_survey_2013}. Probabilistic models have been more successful than deterministic ones, as they provide an estimate about the uncertainty of their approximation which can be incorporated into long-term planning~\cite{deisenroth_gaussian_2015}. For example, local linear models have been used in~\cite{bagnell2001autonomous,ng2006autonomous,levine2014learning}, Gaussian processes (GPs) in~\cite{ko2007gaussian,deisenroth_gaussian_2015,kupcsik2014model} and least-squares conditional density estimation in~\cite{tangkaratt2014model}.

Early examples of such model-based PS include applications on helicopter hovering~\cite{bagnell2001autonomous,ng2006autonomous} and blimp control~\cite{ko2007gaussian}. These works employ the PEGASUS algorithm which can transform a stochastic Markov Decision Process (MDP) or partially-observable MDP (POMDP) into a deterministic POMDP~\cite{ng2000pegasus}.
It does so by fixing in advance the sequence of random numbers associated with the state transitions. This simple modification significantly reduces the time needed to optimize the policy, as it removes the noise from the evaluation of an initially noisy objective function.

Both the model-based PGPE~\cite{tangkaratt2014model} and the PILCO~\cite{deisenroth_gaussian_2015} algorithm use gradient-based policy updates. Rather than using Monte Carlo sampling, as in model-based PGPE, PILCO performs deterministic approximate inference by explicitly incorporating the model uncertainty into long-term predictions. This procedure is done by approximating the probability distribution over trajectories with a Gaussian that has the same mean and covariance (moment matching). The gradient of the expected return is then computed analytically with respect to the policy parameters. This makes PILCO dependent on differentiable reward and policy functions.


Gradient-free methods, such as the Model-Based Relative Entropy PS (M-REPS)~\cite{kupcsik2014model} and the Model-Based Guided PS (M-GPS)~\cite{levine2014learning}, do not have these requirements. Both algorithms place a KL-divergence constraint on the cost function to bound the distance between the old trajectory distribution and the newly estimated one at each policy improvement step. This constraint limits the information loss of the updates~\cite{peters2008reinforcement}. M-GPS turns the policy optimization problem into a supervised learning one, allowing the use of high-dimensional policy representations such as deep neural networks. However, M-GPS makes strong assumptions about the task at hand, by assuming that time-varying Gaussians can approximate the local dynamics. In contrast, M-REPS uses GPs for model learning, and the REPS algorithm (which can be seen as a BBO) for policy search.

Overall, the current consensus~\cite{deisenroth_survey_2013} is that (1) model-based algorithms are more data-efficient than direct PS, (2) in model-based PS, it is crucial to account for potential model errors during policy learning, and (3) deterministic approximate inference and analytic computation of policy gradients is required to make model-based PS computationally tractable. In this paper, we focus on the latter and explore a parallel BBO algorithm for policy optimization.

\section{Problem Formulation}

\noindent We consider dynamical systems of the form:
\begin{align}
 \mathbf{x}_{t+1} = \mathbf{x}_t + f(\mathbf{x}_t,\mathbf{u}_t) + \mathbf{w}
\end{align}
with continuous-valued states $\mathbf{x}\in\mathbb{R}^E$ and controls $\mathbf{u}\in\mathbb{R}^F$, i.i.d. Gaussian system noise $\mathbf{w}$, and unknown transition dynamics $f$.

Our objective is to find a deterministic \textit{policy} $\pi$, $\mathbf{u} = \pi(\mathbf{x}|\boldsymbol{\theta})$, which maximizes the \textit{expected long-term reward} when following policy $\pi$ for $T$ time steps:
\begin{align}
  \label{eq:reward_j}
  J(\boldsymbol{\theta}) = \mathbb{E} \Bigg[\sum_{t=1}^{T}r(\mathbf{x}_t) \Big| \boldsymbol{\theta} \Bigg]
\end{align}
where $r(\mathbf{x}_t)$ is the immediate reward of being in state $\mathbf{x}$ at time $t$.
We assume that $\pi$ is a function parameterized by $\boldsymbol{\theta}\in\mathbb{R}^{\Theta}$ and that the immediate reward function $r(\mathbf{x})\in\mathbb{R}$ might be unknown to the learning algorithm.
%
%

\section{Approach}

\subsection{Learning dynamics model with Gaussian processes}
We would like to have a model $\hat{f}$ that approximates as accurately as possible the unknown dynamics $f$ of our systems and provides uncertainty information.
We rely on Gaussian processes (GPs) to do so. A GP is an extension of multivariate Gaussian distribution to an infinite-dimension stochastic process for which any finite combination of dimensions will be a Gaussian distribution~\cite{rasmussen2006gaussian}. 

As inputs, we use tuples made of the state vector $\mathbf{x}_t$ and the action vector $\mathbf{u}_t$, that is, $\mathbf{\tilde{x}}_t = (\mathbf{x}_t,\mathbf{u}_t)\in\mathbb{R}^{E+F}$;
as training targets, we use the difference between the current state vector and the next one: $\mathbf{\Delta}_{\mathbf{x}_t} = \mathbf{x}_{t+1}-\mathbf{x}_t\in\mathbb{R}^E$. We use $E$ independent GPs to model each dimension of the difference vector $\mathbf{\Delta}_{\mathbf{x}_t}$.
For each dimension $d=1\dots E$ of $\mathbf{\Delta}_{\mathbf{x}_t}$, the GP is computed as ($k_{\hat{f}_d}$ is the kernel function):
\begin{align}
  \hat{f}_d(\mathbf{\tilde{x}})\sim\mathcal{GP}(\mu_{\hat{f}_d}(\mathbf{\tilde{x}}),k_{\hat{f}_d}(\mathbf{\tilde{x}},\mathbf{\tilde{x}}'))
\end{align}
Assuming $D^d_{1:t} = \{f_d(\mathbf{\tilde{x}}_1),...,f_d(\mathbf{\tilde{x}}_t)\}$ is a set of observations, we can query the GP at a new input point $\mathbf{\tilde{x}}_*$:
\begin{align}
  \label{eq:gp}
  p(\hat{f}_d(\mathbf{\tilde{x}}_*)|D^d_{1:t},\mathbf{\tilde{x}}_*) = \mathcal{N}(\mu_{\hat{f}_d}(\mathbf{\tilde{x}}_*),\sigma_{\hat{f}_d}^{2}(\mathbf{\tilde{x}}_*))
\end{align}
The mean and variance predictions of this GP are computed using a kernel vector $\pmb{k}_{\hat{f}_d} = k(D^d_{1:t},\mathbf{\tilde{x}}_*)$, and a kernel matrix $K_{\hat{f}_d}$, with entries $K_{\hat{f}_d}^{ij} = k_{\hat{f}_d}(\mathbf{\tilde{x}}_i,\mathbf{\tilde{x}}_j)$:
\begin{align}
  \label{eq:gp_detail}
  &\mu_{\hat{f}_d}(\mathbf{\tilde{x}}_*) = \pmb{k}_{\hat{f}_d}^{T}K_{\hat{f}_d}^{-1}D^d_{1:t}\nonumber\\
  &\sigma_{\hat{f}_d}^{2}(\mathbf{\tilde{x}}_*) = k_{\hat{f}_d}(\mathbf{\tilde{x}}_*,\mathbf{\tilde{x}}_*)-\pmb{k}_{\hat{f}_d}^{T}K_{\hat{f}_d}^{-1}\pmb{k}_{\hat{f}_d}
\end{align}
In this paper, we use the exponential kernel with automatic relevance determination \cite{rasmussen2006gaussian}:
\begin{align}
  \label{eq:sq_exp_ard}
  k_{\hat{f}_d}(\mathbf{\tilde{x}}_p,\mathbf{\tilde{x}}_q) &= \sigma_d^2\text{exp}(-\frac{1}{2}(\mathbf{\tilde{x}}_p-\mathbf{\tilde{x}}_q)^T\boldsymbol{\Lambda}_d^{-1}(\mathbf{\tilde{x}}_p-\mathbf{\tilde{x}}_q))\nonumber\\
  &+ \delta_{pq}\sigma_{n_d}^2
\end{align}

where $\delta_{pq}$ equals to 1 when $p=q$ and 0 otherwise, and $[\boldsymbol{\Lambda}_d, \sigma_d^2, \sigma_{n_d}^2]$ is the vector of hyper-parameters of the kernel (length scales for each dimension of the observations, signal variance and noise)
found through Maximum Likelihood Estimation~\cite{rasmussen2006gaussian}. 
We use the limbo C++11 library for GP regression~\cite{cully2016limbo}.

\subsection{Learning the immediate reward function with a GP}

Similarly to the dynamical model, we use a GP to learn the immediate reward function, which associates a reward $r(\mathbf{x}) \in \mathbb{R}$ to each state $\mathbf{x}$:
\begin{align}
  \label{eq:gp_rew}
  \hat{r}(\mathbf{x})\sim\mathcal{GP}(\mu_r(\mathbf{x}),k_r(\mathbf{x},\mathbf{x}'))
\end{align}
The GP predictions are calculated similarly to Eq.~\ref{eq:gp} and~\ref{eq:gp_detail}.

\begin{figure*}
  \centering
\includegraphics[width=0.99\linewidth]{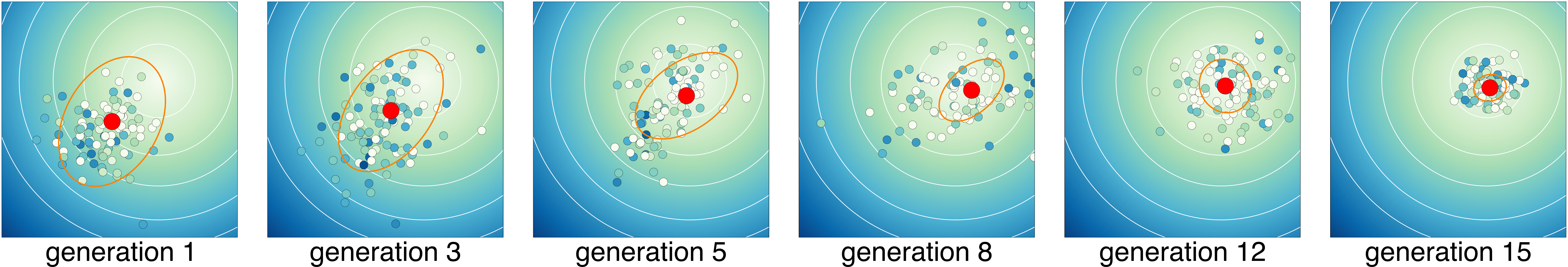}
\vspace{-1em}
\caption{\label{fig:cmaes}Illustration of an actual run with CMA-ES on a simple 2-D, noisy problem. The performance landscape is pictured on the background. Each colored disk is a candidate from the population (the color represents its performance with the same color scale as the landscape in the background). If the color of a disk is the same as the background, then noise did not change the performance. The mean $m_k$ of the best $\mu$ candidates is pictured as a red disk, and the covariance of the $\mu$ best individuals as an orange ellipse. In only a few generations, and in spite of the noise, CMA-ES identifies the optimum of the function. Please note that in this example we use a bigger population than in our work, as advised by the authors of CMA-ES~\cite{hansen2001completely}.}
\vspace{-1.5em}
\end{figure*}

\subsection{Policy Evaluation}
Our goal is to maximize the expected cumulative reward (Eq.~\ref{eq:reward_j}), which requires predicting the state evolution given an uncertain transition model and an uncertain reward model. To do so in a deterministic way\footnote{Additionally, PILCO requires the reward function to be known a priori.}, PILCO proposes to approximate the distribution of state $\mathbf{x}_{t+1}$ given the distribution of state $\mathbf{x}_t$ and the action $\mathbf{u}_t$ using moment matching~\cite{deisenroth_gaussian_2015}, and then propagates from state to state until reaching the end of the episode; however, this sequential approach accumulates errors over time, is not easy to parallelize, and is computationally expensive. As an alternative, we can compute a Monte Carlo approximation of the final distribution: at each step we sample a new state according to the GP of the model and its reward according to the reward model, query the policy to choose the actions, and use this new state to sample the next state. By performing this process many times, we can get a good estimate of the expected cumulative reward, but many samples are needed to obtain a good estimate~\cite{kupcsik2014model}.

Here we adopt a different approach. Like in Monte Carlo estimation, we propagate from state to state by sampling according to the models. However, we consider that each of these rollouts is a measurement of a function $G(\boldsymbol{\theta})$ that is the actual function $J(\boldsymbol{\theta})$ perturbed by a noise $N(\boldsymbol{\theta})$:
\begin{align}
  \label{eq:objective_g}
  G(\boldsymbol{\theta}) &= J(\boldsymbol{\theta})  + N(\boldsymbol{\theta})\nonumber\\
  &= \sum_{t=1}^{T}\hat{r}(\mathbf{x}_{t-1}+\hat{f}(\mathbf{x}_{t-1}, \mathbf{u}_{t-1}))
\end{align}
where $\hat{f}(\mathbf{x}_{t-1}, \mathbf{u}_{t-1})\sim\mathcal{N}(\mu_{\hat{f}}(\mathbf{\tilde{x}}_{t-1}),\Sigma_{\hat{f}}(\mathbf{\tilde{x}}_{t-1}))$ is a realization of a normally distributed random vector according to Eq.~\ref{eq:gp},
$\hat{r}(\mathbf{x})\sim\mathcal{N}(\mu_r(\mathbf{x}),\sigma_r^2(\mathbf{x}))$ is a realization of a normally distributed random value according to Eq.~\ref{eq:gp_rew} and
$\mathbf{u}_{t-1} = \pi(\mathbf{x}_{t-1}|\boldsymbol{\theta})$.
We would like to maximize its expectation:
\begin{align}
  \mathbb{E}\Big[G(\boldsymbol{\theta})\Big]
    &= \mathbb{E}\Big[J(\boldsymbol{\theta})  + N(\boldsymbol{\theta})\Big]\nonumber\\
    & = \mathbb{E}\Big[J(\boldsymbol{\theta})\Big] + \mathbb{E}\Big[N(\boldsymbol{\theta})\Big]
  \end{align}
  And since $\forall x$ $\mathbb{E}\Big[\mathbb{E}[x]\Big] = \mathbb{E}[x]$:
\vspace{-0.5em}
\begin{equation}
  \label{eq:policy_eval}
  \mathbb{E}\Big[G(\boldsymbol{\theta})\Big]   = J(\boldsymbol{\theta}) + \mathbb{E}\Big[N(\boldsymbol{\theta})\Big]
\end{equation}
We assume that $\mathbb{E}[N(\boldsymbol{\theta})] = 0$ for all $\boldsymbol{\theta}\in\mathbb{R}^{\Theta}$ and therefore maximizing $\mathbb{E}[G(\boldsymbol{\theta})]$ is equivalent to maximizing $J(\boldsymbol{\theta})$.

\subsection{Policy search}
\label{sec:cmaes_uncertain}
Seeing the maximization of $J(\boldsymbol{\theta})$ as the optimization of a noisy function allows us to maximize it without computing or estimating it explicitly: we only use the noisy measurements in the optimization algorithm. To do so, we build on all the work about noisy function optimization~\cite{jin2005evolutionary,hansen2009method}, and especially on CMA-ES, one of the most successful black-box optimizer for noisy functions~\cite{hansen2001completely}.
CMA-ES (Fig. \ref{fig:cmaes}) performs four steps at each generation $k$:
\begin{itemize}
  \item[(1)] sample $\lambda$ new candidates according to a multi-variate Gaussian distribution of mean $m_k$ and covariance $\sigma_k^2 C_k$, that is, $\boldsymbol{\theta}_i \sim \mathcal{N}(m_k, \sigma_k^2 C_k)$ for $i \in 1, \cdots, \lambda$;
  \item[(2)] rank the $\lambda$ sampled candidates based on their (noisy) performance $G(\boldsymbol{\theta}_i)$;
  \item[(3)] compute $m_{k+1}$ by averaging the $\mu$ best candidates: $m_{k+1} = \frac{1}{\mu}\sum_{i=1}^\mu\boldsymbol{\theta}_i$;
  \item[(4)] update the covariance matrix to reflect the distribution of the $\mu$ best candidates.
\end{itemize}
Overall, these steps are only marginally impacted by noise in the performance function, as confirmed by empirical experiments with noisy functions~\cite{jin2005evolutionary}. More precisely, the only decision that matters is whether a solution belongs to the $\mu$ best ones (step 2), that is, a precise ranking is not needed and errors can only happen at the boundaries between the low-performing and high-performing solutions. In addition, if a candidate is misclassified because of the noise, the impact of this error will be smoothed out by the average when computing $m_{k+1}$ (step 3). One can also observe that because CMA-ES samples several solutions around a mean $m_k$, it performs many evaluations of similar parameters, which are then averaged: this \emph{implicit averaging}~\cite{jin2005evolutionary,miller1996genetic} has many similarities with re-evaluating noisy solutions to estimate their expectation.

Modern implementations of CMA-ES add several refinements to compute the covariance matrix, to take into account successive steps, and to restart the process with more exploration when it reaches an optimum. In this work, we use BIPOP-CMA-ES with restarts~\cite{hansen2009benchmarking}, which is one of best CMA-ES variants on benchmarks with both noiseless and noisy functions~\cite{hansen2009benchmarking,hansen2009benchmarking_noisy}.

On top of these refinements, we follow the strategy proposed by Hansen et al.~\cite{hansen2009method} to improve the behavior of CMA-ES with noisy functions (called UH-CMA-ES). The starting idea is that uncertainty is a problem for a rank-based algorithm if and only if, for two potential candidates $\boldsymbol{\theta}_1$ and $\boldsymbol{\theta}_2$ the variation due to $N(\boldsymbol{\theta}_1)$ and $N(\boldsymbol{\theta}_2)$ exceeds the difference $\lvert J(\boldsymbol{\theta}_1)-J(\boldsymbol{\theta}_2)\rvert$ and thus their ordering is changed.
If the variation tends to exceed this difference, we cannot conclude only from two measurements $G(\boldsymbol{\theta}_1)$, $G(\boldsymbol{\theta}_2)$, whether $J(\boldsymbol{\theta}_1)>J(\boldsymbol{\theta}_2)$ or $J(\boldsymbol{\theta}_1)<J(\boldsymbol{\theta}_2)$ holds.
If we view $\lvert J(\boldsymbol{\theta}_1)-J(\boldsymbol{\theta}_2)\rvert$ as the signal and the variations due to $N(\boldsymbol{\theta})$ as noise, then it follows that one way to improve the quality of the ranking without re-evaluating solutions many times (which would reduce noise) is to increase the signal.

We therefore implement the following strategy: (1) at each generation, we quantify the uncertainty of the ranking by re-evaluating $\lambda_{\text{reev}}<\lambda$ randomly selected candidates from the population and count the number of rank changes (see~\cite{hansen2009method} for a detailed description of uncertainty quantification), (2) if the uncertainty is above a user-defined threshold, then we increase the variance of the population ($\sigma_k$ in step 1 of CMA-ES). In addition of reducing the uncertainty of the ranking when needed, this strategy has an interesting consequence: in uncertain search-space regions, CMA-ES moves faster (it makes bigger steps), which means that the algorithm favors regions that are more certain (when they are as promising as uncertain regions) and is not ``trapped'' in uncertain regions. We use a modified version of the \emph{libcmaes} C++11 library\footnote{\scriptsize\url{https://github.com/beniz/libcmaes}}.
\subsection{Black-box Data-efficient Robot Policy Search}

Putting everything together, we get the Black-DROPS algorithm (Alg.~\ref{algo:bdrops}). Firstly, $N_R$ random episodes of $T$ time steps are conducted on the robot (Alg.~\ref{algo:bdrops}: lines 4-12). In the learning loop, first we learn a probabilistic model of the dynamics and a model of the reward function, and then we optimize $\mathbb{E}\big[G(\boldsymbol{\theta})\big]$ given this learned models using BIPOP-CMAES with uncertainty handling (Alg.~\ref{algo:bdrops}: lines 14-16). Lastly, the best policy $\pi_{\boldsymbol{\theta}^*}$ is executed on the robot, more data is collected and the main loop continues until the task is learned.

\begin{algorithm}
  \footnotesize
  \caption{Black-DROPS}
  \label{algo:bdrops}
  \begin{algorithmic}[1]
    \Procedure{Black-DROPS}{}
      \State Define policy $\pi: \mathbf{x}\times\boldsymbol{\theta}\to\mathbf{u}$
      \State $D = \emptyset$
      \For {$i=1\to N_R$} \Comment{$N_R$ random episodes}
        \State Set robot to initial state $\mathbf{x}_0$
        \For {$j=0\to T-1$} \Comment{perform the episode}
          \State $\mathbf{u}_j$ = random\_action()
          \State $\mathbf{x}_{j+1},r(\mathbf{x}_{j+1})$ = execute\_on\_robot($\mathbf{u}_j$)
          \State $D = D \cup \{\mathbf{\tilde{x}}_j\to\mathbf{\Delta}_{\mathbf{x}_j}\}$
          \State $R = R \cup \{\mathbf{x}_{j+1}\to r(\mathbf{x}_{j+1})\}$
        \EndFor

      \EndFor

      \While{$task \ne solved$}
        \State Model learning: train $E$ GPs given data $D$
        \State Reward learning: train 1 GP given data $R$
        \State $\boldsymbol{\theta}^* = \argmax_{\boldsymbol{\theta}}\mathbb{E}\Big[G(\boldsymbol{\theta})\Big]$ using BIPOP-CMA-ES \Comment{Sec.~\ref{sec:cmaes_uncertain}}
        \State Set robot to initial state $\mathbf{x}_0$
        \For {$j=0\to T-1$} \Comment{perform the episode}
          \State $\mathbf{u}_j = \pi(\mathbf{x}_j|\boldsymbol{\theta}^*)$
          \State $\mathbf{x}_{j+1},r(\mathbf{x}_{j+1})$ = execute\_on\_robot($\mathbf{u}_j$)
          \State $D = D \cup \{\mathbf{\tilde{x}}_j\to\mathbf{\Delta}_{\mathbf{x}_j}\}$
          \State $R = R \cup \{\mathbf{x}_{j+1}\to r(\mathbf{x}_{j+1})\}$
        \EndFor
      \EndWhile

    \EndProcedure
  \end{algorithmic}
\end{algorithm}

\section{Experimental Setup}

\subsection{Policy Representations}
To highlight the flexibility of Black-DROPS, we use a GP-based policy~\cite{deisenroth_gaussian_2015} and a feed-forward neural network-based one. Any other parameterized policy can be used (e.g., dynamic movement primitives).

\subsubsection{GP Policy}
\label{sec:gp_policy}
If we only consider the mean, a Gaussian process can be used to map states to actions, that is, to define a policy:
\begin{align}
  \pi(\mathbf{x}) = \mathbf{u}_{\text{max}}\kappa(\mu(\mathbf{x})) = \mathbf{u}_{\text{max}}\kappa(\pmb{k}^{T}(K+\sigma_{n}^{2}I)^{-1}\mathbf{x})
\end{align}
where $\mathbf{u}_{\text{max}}$ is the maximum value of $\mathbf{u}$ (different for each action dimension),
$\kappa$ is a squashing function like the one used in~\cite{deisenroth_gaussian_2015}, $\mathbf{x}$ is the input state vector to the policy, $K$ is the covariance matrix and its elements are computed using the exponential kernel with automatic relevance determination as in Eq.~\ref{eq:sq_exp_ard}. Here, we set signal noise, $\sigma_n^2 = 0.01$.
The vector $[\boldsymbol{\Lambda}, \sigma_f^2]$ and the pseudo-observations (\emph{inputs} \& \emph{targets} when learning the GP) constitute the parameters of the policy.

\subsubsection{Neural Network Policy}
\label{sec:nn_policy}
The network function of the $i^{th}$ layer of the network is given by $\mathbf{y}_i = \phi_i(\mathbf{W}_i\mathbf{y}_{i-1}+\mathbf{b}_i)$,
where $\mathbf{W}_i$ and $\mathbf{b}_i$ are the weight matrix and bias vector, $\mathbf{y}_{i-1}$ and $\mathbf{y}_i$ are the input and output vector and $\phi_i$ is the activation function.
Throughout the paper, we use configurations with one hidden layer and the hyperbolic tangent as the activation function $\phi$ for all the layers, leading to:
\begin{align}
  \pi(\mathbf{x}) = \mathbf{u}_{\text{max}}\mathbf{y}_1 = \mathbf{u}_{\text{max}}\phi(\mathbf{W}_1\mathbf{y}_{0}+\mathbf{b}_1)&\nonumber\\
  \text{and }\mathbf{y}_0 = \phi(\mathbf{W}_0\mathbf{x}+\mathbf{b}_0)&
\end{align}
%
%

\subsection{Metrics}

\subsubsection{Reward as interaction time increases}

This metric assesses the quality of the solutions and the data-efficiency of each algorithm.

\subsubsection{Speed-up when more cores are available}

This metric assesses how well each algorithm scales as the available hardware resources increase, independently of the particular implementation (e.g., MATLAB vs C++).

\subsection{Remarks}

We evaluate Black-DROPS on the pendulum and cart-pole tasks and compare it to PILCO using 120 replicates over different CPU configurations. As an additional baseline, we evaluate a variant of our approach using deterministic GP models of the dynamics (i.e., using only the mean of the GPs) to quantify the importance of considering the uncertainty (variance) of the model in policy optimization. For Black-DROPS and the baseline we use two different policies: a neural network policy (with one hidden layer and 10 hidden units) and a GP policy (with 10 and 20 pseudo-observations for the pendulum and the cart-pole task respectively).
For PILCO we used only the GP policy with the same parameters as for the other algorithms.

We additionally evaluate Black-DROPS on a 4-DOF arm task to validate that it can be used with more complex and interesting robots, 
that it can be used when the reward function is unknown, and that it works on a real robotic platform.  We use only the neural network policy for this task, as it performed better in the simpler benchmarks.

For all the tasks, an episode corresponds to applying the same policy for a duration of $4\,s$ and the sampling/control rate is $10Hz$.
The source code of the experiments can be found at {\footnotesize\url{https://github.com/resibots/blackdrops}}.

\section{Results}

\begin{figure}
  \centering
  \includegraphics[width=\linewidth]{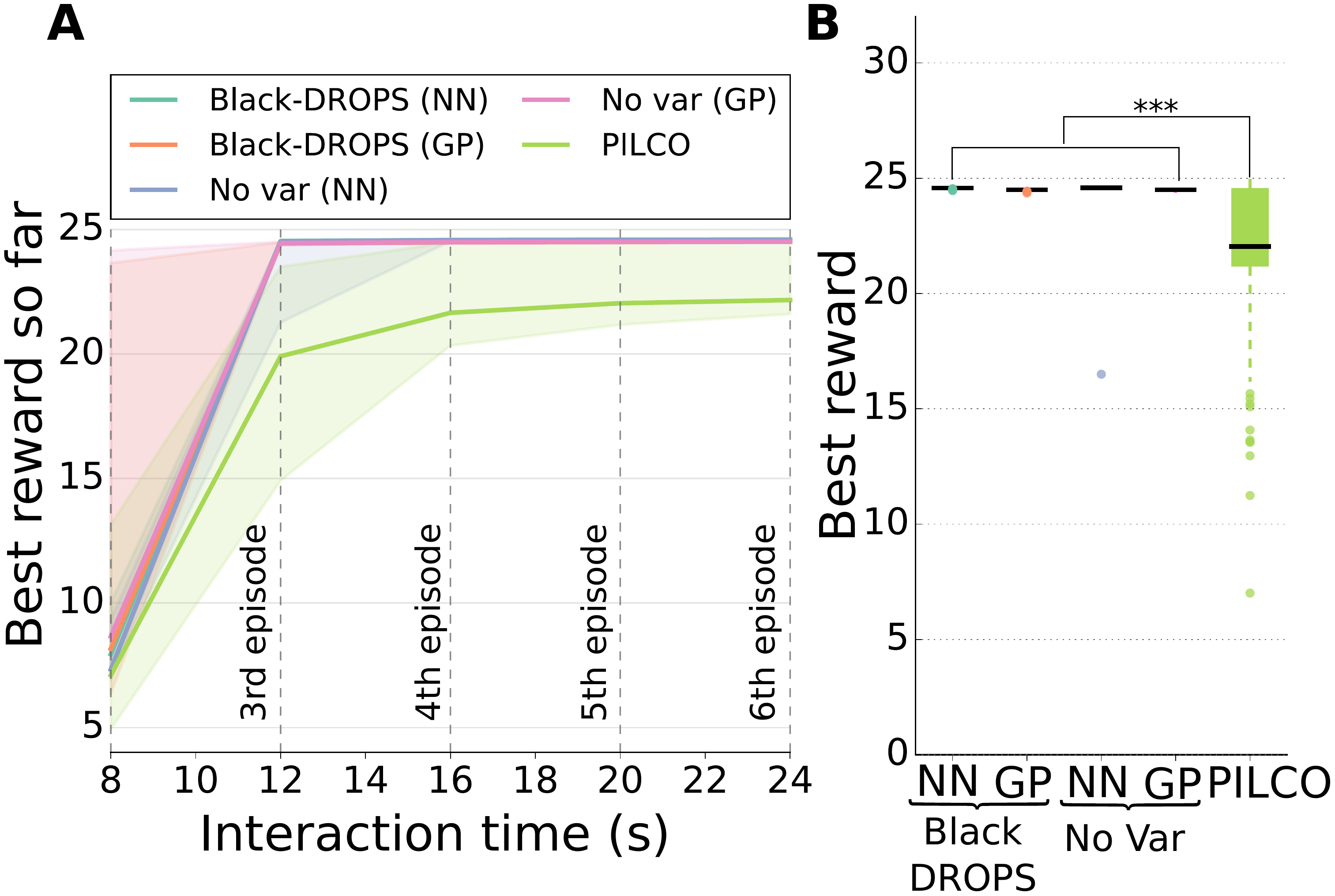}
  \vspace{-1.2em}
  \caption{Results for the pendulum task (120 replicates): (\textbf{A}) Best reward found per episode. The lines are median values and the shaded regions the $25^{th}$ and $75^{th}$ percentiles. Black-DROPS converges to higher quality solutions in fewer episodes than PILCO and has considerably less variance. (\textbf{B}) Best reward after 4 episodes. The box plots show the median (black line) and the interquartile range ($25^{th}$ and $75^{th}$ percentiles); the whiskers extend to the most extreme data points not considered outliers, and outliers are plotted individually. Our approach outperforms PILCO in the quality of the controllers found. The number of stars indicates that the p-value of the Mann-Whitney U test is less than $0.05$, $0.01$, $0.001$ and $0.0001$ respectively.}
  \label{fig:pend_rewards}
  \vspace{-1.5em}
\end{figure}

\subsection{Task 1: Inverted Pendulum}

This simulated system consists of a freely swinging pendulum with mass $m=1\,kg$ and length $l=1\,m$. 
The objective is to learn a controller to swing the pendulum up and to balance it in the inverted position applying a torque.
\begin{itemize}
  \item \textbf{State:} $\mathbf{x}_{pend} = [\dot{\theta}, \theta]\in\mathbb{R}^2$, $\mathbf{x}_0 = [0,0]$.
  \item \textbf{Actions:} $\mathbf{u}_{pend} = u_{pend}\in\mathbb{R}$, $-2.5\leq u_{pend}\leq 2.5\,N$.
  \item To avoid angle discontinuities, we transform the input of the GPs, the reward function, and the policy to be:
  \begin{equation*}
    \mathbf{x}_{input} = [\dot{\theta}, \text{cos}(\theta), \text{sin}(\theta)]\in\mathbb{R}^3
  \end{equation*}
  The MATLAB implementation of PILCO uses this transformation by default\footnote{http://mlg.eng.cam.ac.uk/pilco/}.
  \item \textbf{Reward:} We use the same reward function as PILCO\footnote{PILCO uses a cost function, but it is straightforward to transform it in a reward function.}. This is a saturating distance-based reward function:
  \begin{equation}
    \label{eq:reward_function}
    r(\mathbf{x}) = \text{exp}(-\frac{1}{2\sigma_c^2}(\mathbf{x}-\mathbf{x}_{*})^T\mathbf{Q}(\mathbf{x}-\mathbf{x}_{*}))
  \end{equation}

  where $\sigma_c$ controls the width of the reward function, $\mathbf{Q}$ is a weight matrix, $\mathbf{x}_{*}$ is the target state and $r(\mathbf{x})\in[0,1]$.
  We set $\mathbf{x}_{*} = [*, \text{cos}(\pi), \text{sin}(\pi)]$, $\sigma_c = 0.25$ and $\mathbf{Q}$ to ignore the angular velocity $\dot{\theta}$ of the pendulum.
\end{itemize}

\begin{figure}
  \centering
  \includegraphics[width=\linewidth]{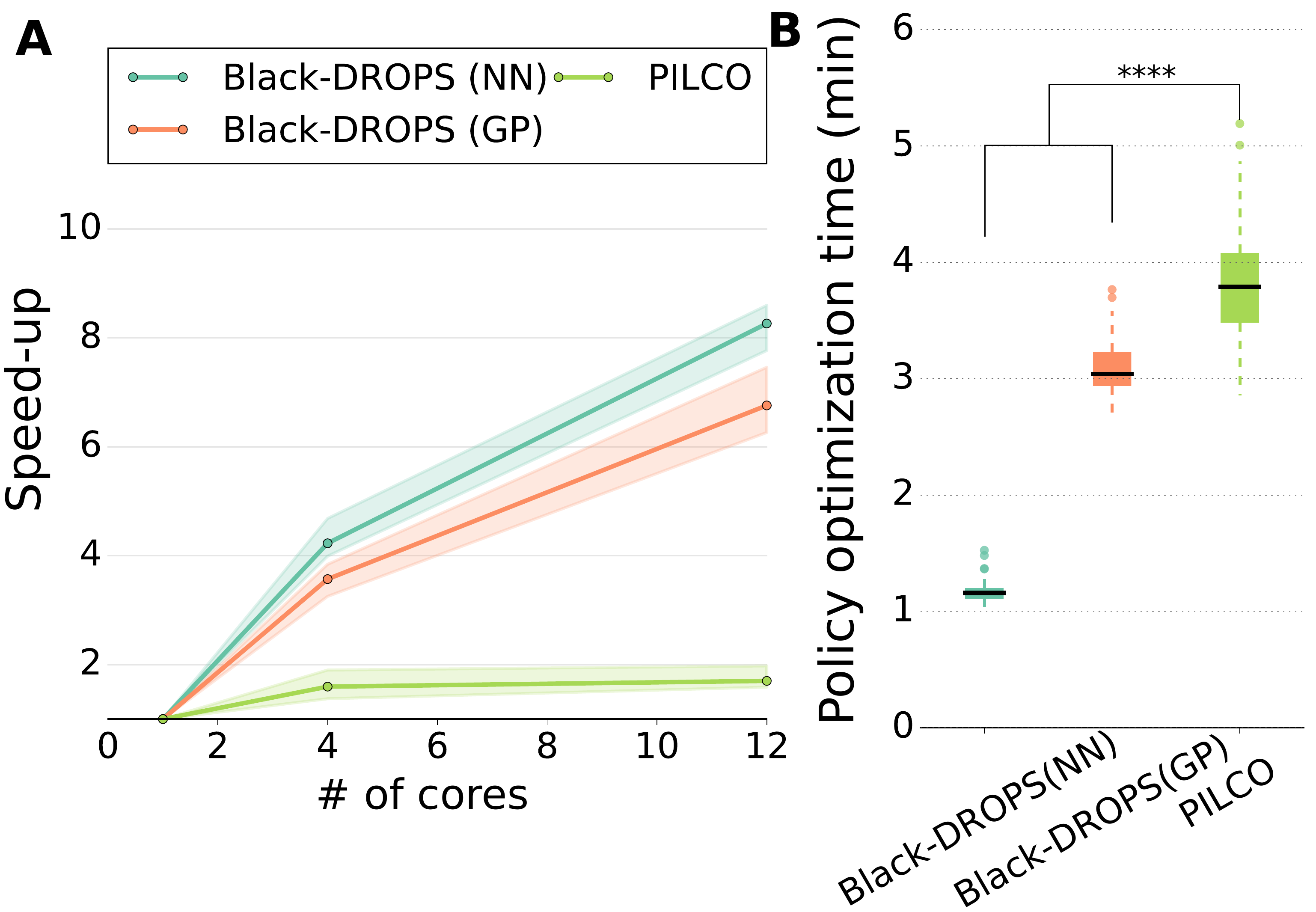}
  \vspace{-1.2em}
  \caption{Timing for the the pendulum task: (\textbf{A}) Speed-up (for total policy optimization time after 4 episodes) achieved when using multiple cores. The lines are median values over 30 runs and the shaded regions the $25^{th}$ and $75^{th}$ percentiles. As more cores are being used, Black-DROPS greatly benefits from it and has up to $8$x speed-up when 12 cores are used. (\textbf{B}) Total policy optimization time after 4 episodes when 12 cores are available.}
  \label{fig:pend_results}
  \vspace{-1em}
\end{figure}

In this task, both Black-DROPS and PILCO solve the task in about 3 episodes ($12\,s$ of interaction time --- including the random episode, Fig.~\ref{fig:pend_rewards}\textbf{A}), but Black-DROPS finds higher-performing policies (Fig.~\ref{fig:pend_rewards}\textbf{A-B}), with both the neural network and the GP policy. When the number of cores is increased, the computation time required by Black-DROPS decreases almost linearly (Fig.~\ref{fig:pend_results}\textbf{A}), whereas PILCO only slightly benefits from having more than 4 cores (as PILCO is not a parallel algorithm, we think that the improvement between 1 and 4 cores stems from MATLAB's ability to parallelize some linear algebra operations).
With more than 8 cores, Black-DROPS outperforms PILCO in computation speed and can be from $1.25$ to $3.3$ times faster when 12 cores are available\footnote{While some of the runtime differences can stem from the language used (e.g., C++ being faster than MATLAB or MATLAB being faster at matrix computations), what matters is that a parallel algorithm with enough CPUs can eventually outperform a sequential gradient-based approach.}
(Fig.~\ref{fig:pend_results}\textbf{B}). In addition, given a budget of 15 episodes, Black-DROPS succeeds more often than PILCO in finding a working policy (Table~\ref{tab:pend_success}): Black-DROPS always solves the task whereas PILCO fails once in ten runs.

Surprisingly, taking into account the uncertainty of the model does not seem to be necessary in this task (the ``No Var'' baselines perform the same as Black-DROPS, see Fig.~\ref{fig:pend_rewards}). This result most probably stems from the fact that the dynamics of the system are simple enough for the GPs to model almost perfectly with one or two episodes.
%
%
%
%
\begin{figure}
  \centering
  \includegraphics[width=\linewidth]{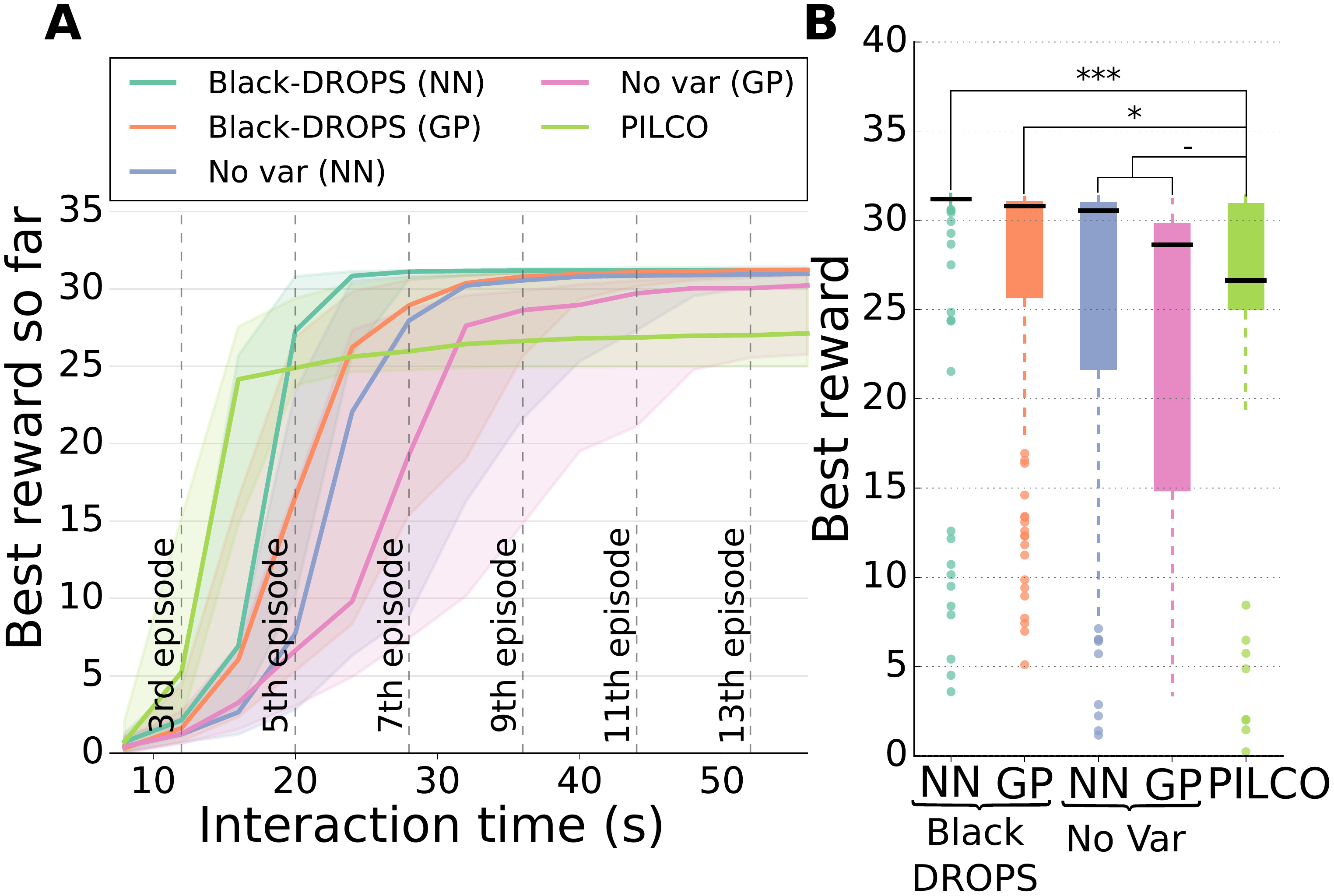}
  \vspace{-1.2em}
  \caption{Results for the cart-pole task (120 replicates): (\textbf{A}) Best reward found per episode. Black-DROPS converges to higher quality solutions in about the same number of episodes as PILCO and has less variance. (\textbf{B}) Best reward after 8 episodes. Our approach outperforms PILCO in the quality of the controllers found. See Fig.~\ref{fig:pend_rewards} for legend.}
  \label{fig:cp_rewards}
  \vspace{-1.5em}
\end{figure}
\begin{table}[!htb]
\def\arraystretch{1.2}
\begin{center}
\caption{}
\label{tab:pend_success}
\scriptsize
\begin{tabulary}{\linewidth}{|l||c|c|}
  \hline
  \multicolumn{1}{|c||}{\textbf{Algorithm}} & \multicolumn{2}{c|}{\textbf{Success Rate}} \\
  \hline
  & \textbf{Pendulum} & \textbf{Cart-pole}\\
  \hline
  Black-DROPS (NN) & $100\%$ & $93.33\%$ \\
  \hline
  Black-DROPS (GP) & $100\%$ & $90.83\%$ \\
  \hline
  No Var (NN) & $100\%$ & $90\%$ \\
  \hline
  No Var (GP) & $100\%$ & $85\%$ \\
  \hline
  PILCO & $89.16\%$ & $92.5\%$ \\
  \hline
\end{tabulary}
\end{center}
\vspace{-2em}
\end{table}

\subsection{Task 2: Cart-pole Swing-Up}

This simulated system consists of a cart
with mass $M=0.5\,kg$ running on a track and a freely swinging pendulum with mass $m=0.5\,kg$ and length $l=0.5\,m$ attached to the cart. The state of the system contains the position of the cart, the velocity of the cart, the angle of the pendulum and the angular velocity of the pendulum. 
The objective is to learn a controller that applies horizontal forces on the cart to swing the pendulum up and balance it in the inverted position in the middle of the track.
\begin{itemize}
  \item \textbf{State:} $\mathbf{x}_{cp} = [\dot{x}, x, \dot{\theta}, \theta]\in\mathbb{R}^4$, $\mathbf{x}_0 = [0,0,0,0]$.
  \item \textbf{Actions:} $\mathbf{u}_{cp} = u_{cp}\in\mathbb{R}$, $-10\leq u_{cp}\leq 10\,N$.
  \item To avoid angle discontinuities, we transform the input of the GPs, the reward, and the policy to be:
  \begin{equation*}
    \mathbf{x}_{input} = [\dot{x}, x, \dot{\theta}, \text{cos}(\theta), \text{sin}(\theta)]\in\mathbb{R}^5
  \end{equation*}
  \item \textbf{Reward:} We set $\mathbf{x}_{*} = [*, 0, *, \text{cos}(\pi), \text{sin}(\pi)]$, $\sigma_c = 0.25$, $\mathbf{Q}$ to ignore $\dot{x}$ and $\dot{\theta}$, and use Eq.~\ref{eq:reward_function}.
\end{itemize}

\begin{figure}
  \centering
  \includegraphics[width=\linewidth]{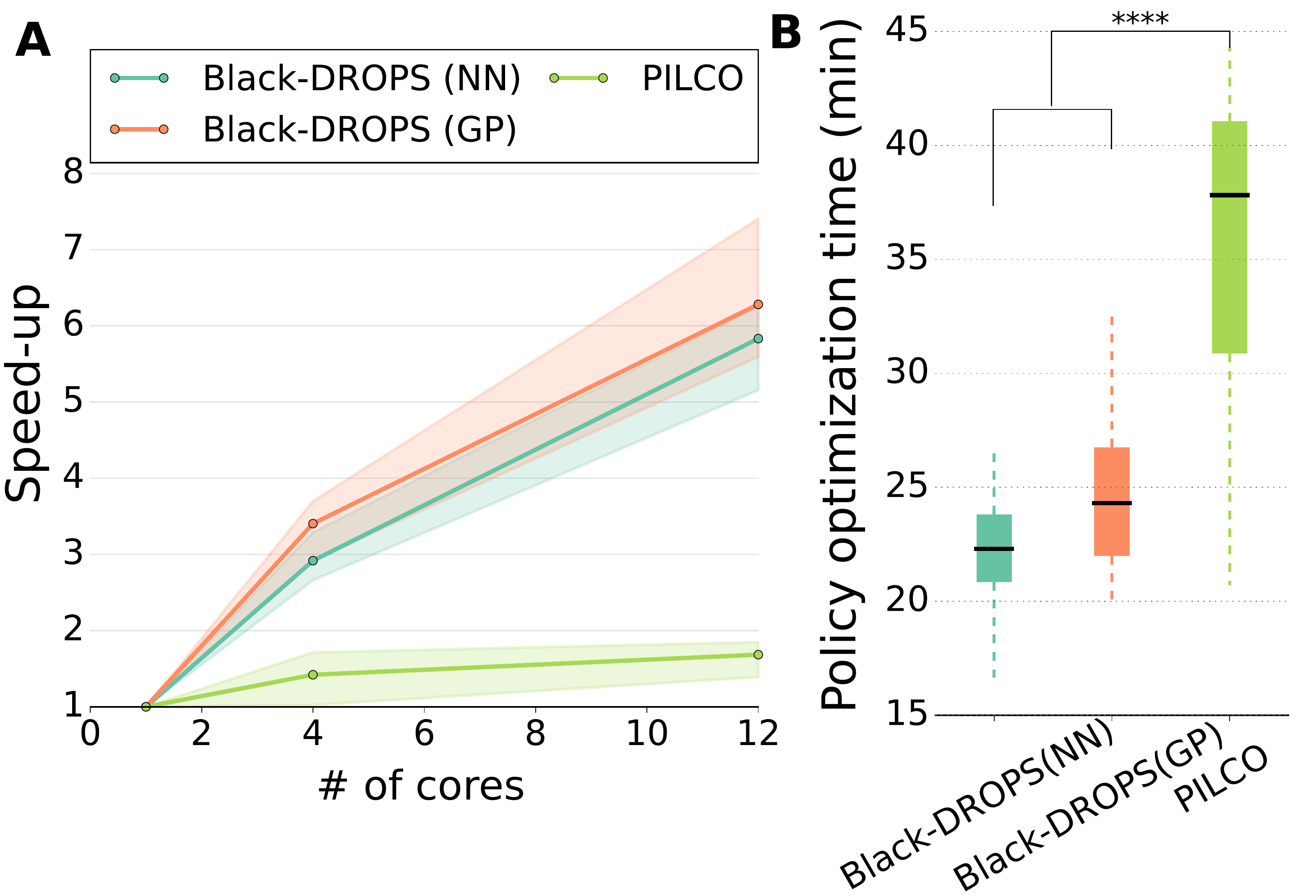}
  \vspace{-1.2em}
  \caption{Timing for the cart-pole task:  (\textbf{A}) Speed-up (for total policy optimization time after 8 episodes) achieved when using multiple cores. As more cores are being used, Black-DROPS greatly benefits from it and has a $6$x speed-up when 12 cores are used. (\textbf{B}) Total policy optimization time after 8 episodes when 12 cores are available. Black-DROPS is around $1.6$x faster than PILCO. See Fig.~\ref{fig:pend_results} for legend and number of replicates.}
  \label{fig:cp_results}
  \vspace{-1.5em}
\end{figure}
\begin{figure}
  \centering
  \includegraphics[width=0.95\linewidth]{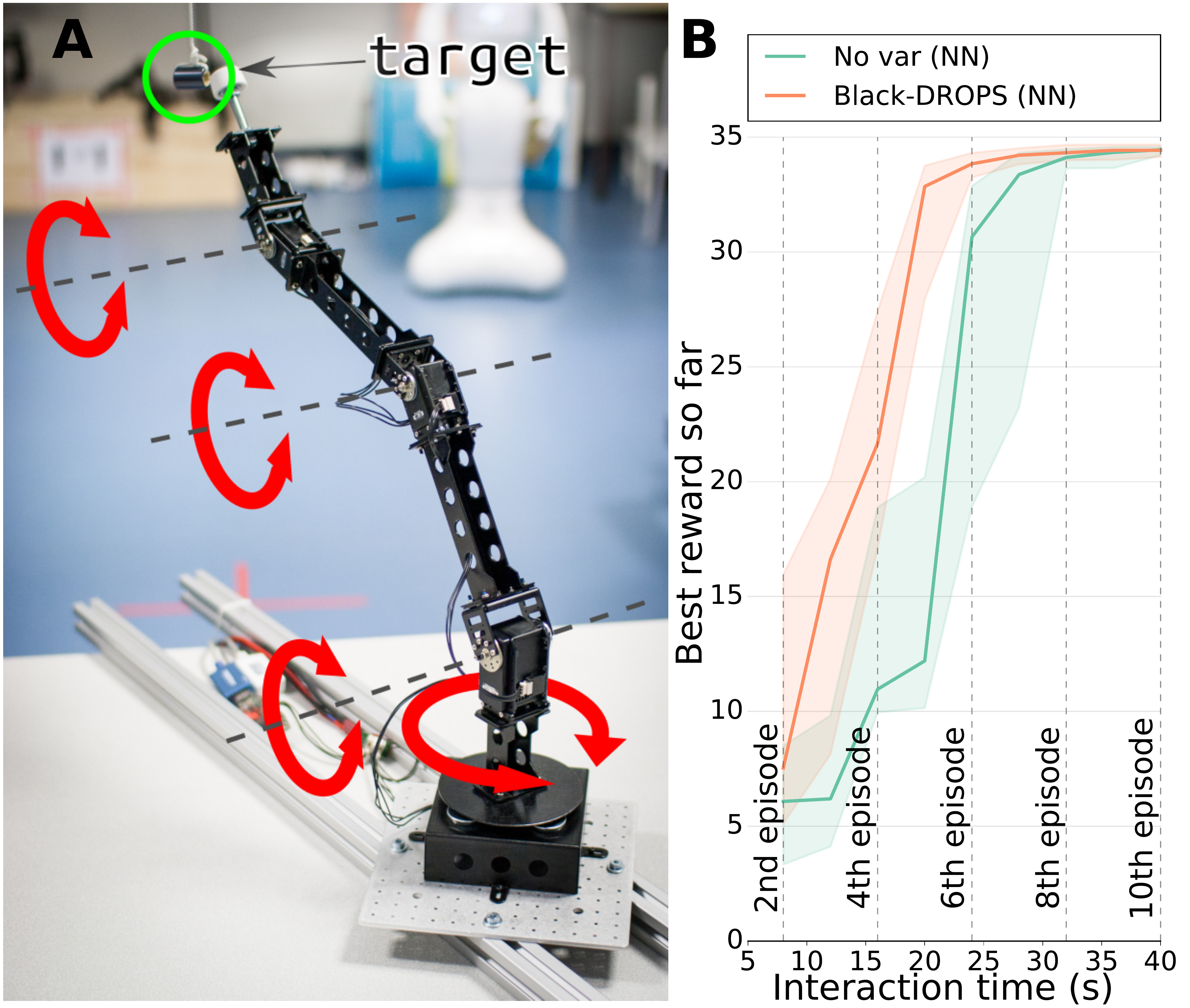}
  \vspace{-1em}
  \caption{Manipulator task (10 replicates for each treatment).}
  \label{fig:manip}
  \vspace{-1.5em}
\end{figure}

The results for the cart-pole are very similar to those obtained with the inverted pendulum (Fig.~\ref{fig:cp_rewards}\textbf{A-B})
: Black-DROPS and PILCO have similar data-efficiency (about 4 episodes or $16\,s$ of interaction time to find a working policy with PILCO, 5 episodes or $20\,s$ with Black-DROPS) but Black-DROPS finds higher-performing policies, most probably because its search algorithm (CMA-ES with restarts) is less local than the gradient-based optimizer used in PILCO.

Using the variance helps more in this task than in the pendulum task: the variants of Black-DROPS without uncertainty handling are less data-efficient and have more variance. However, they are still able to find policies that are higher-performing than those found by PILCO (Fig.~\ref{fig:cp_rewards}\textbf{B}). In terms of success rate, Black-DROPS fails as often as PILCO and, as expected, the variants fail more often than PILCO and Black-DROPS (see Table~\ref{tab:pend_success}).

Similar to the pendulum task, here also Black-DROPS takes advantage of multiple cores to highly speed-up its computation and is $1.6$ times faster than PILCO when 12 cores are available (Fig.~\ref{fig:cp_results}).

\subsection{Task 3: 4-DOF Manipulator}

We applied Black-DROPS on a physical velocity-controlled 4-DOF robotic arm (Fig.~\ref{fig:manip}, 10 replicates). We assume that we can only observe the angles of the joints of the arm 
and that the reward function $r_{arm}$ is initially unknown. 
The arm begins in the up-right position and the objective is to learn a controller so that the end-effector quickly reaches a certain position (shown in Fig.~\ref{fig:manip}\textbf{A}). We compare Black-DROPS with the baseline without variance.
%
\begin{itemize}
  \item \textbf{State:} $\mathbf{x}_{arm} = [q_0, q_1, q_2, q_3]\in\mathbb{R}^4$, $\mathbf{x}_0 = [0,0,0,0]$.
  \item \textbf{Actions:} $\mathbf{u}_{arm} = [v_0, v_1, v_2, v_3]\in\mathbb{R}^4$, where $-1.0\leq v_i\leq 1.0\,rad/s,\quad i=0,1,2,3$.
  \item \textbf{Reward:} The unknown (to the algorithm) reward function has a form similar to Eq.~\ref{eq:reward_function}:
  \begin{equation}
    r_{arm}(\mathbf{x}) = \text{exp}(-\frac{1}{2\sigma_c^2}\lVert\mathbf{p}_{\mathbf{x}}-\mathbf{p}_{*}\rVert)
  \end{equation}
where $\sigma_c = 0.1$, $\mathbf{p}_{\mathbf{x}}$ corresponds to the end-effector position in state $\mathbf{x}$, $\mathbf{p}_{*}$ is the goal position of the end-effector and $r_{arm}(\mathbf{x})\in[0,1]$.
  \item To avoid angle discontinuities, we transform the input to the GPs and the policy to be: \begin{align*}\mathbf{x}_{input} = [\text{cos}(q_0), \text{sin}(q_0), \text{cos}(q_1), \text{sin}(q_1), \text{cos}(q_2),&\\\text{sin}(q_2), \text{cos}(q_3), \text{sin}(q_3)]\in\mathbb{R}^8&\end{align*}
\end{itemize}
The results show that Black-DROPS is able to find a working policy within 5 episodes (including the initial random one) and outperforms the baseline which needs around 6 episodes (Fig.~\ref{fig:manip}\textbf{B}). Black-DROPS, also, shows less variance and converges to high quality controllers faster (6 episodes vs 8-9). A video of a typical run is available as supplementary material (also at {\scriptsize\url{https://youtu.be/kTEyYiIFGPM}}).

\section{Conclusion and discussion}
%
Black-DROPS lifts several constraints imposed by analytical approaches (reward and policy types) while being competitive in terms of data-efficiency and computation time. In three different tasks, it achieved similar results as the state-of-the-art (PILCO) while being faster when multiple cores are used. We expect that the ability of Black-DROPS to scale with the number of cores will be even more beneficial on future computers with more cores and/or with GPUs.

Using the variance in the optimization is one of the key components to learn with as little interaction time as possible. However, the learned dynamics models are only confident in areas of the state space previously visited and thus could drive the optimization into local optima when multiple and diverse solutions exist. In future work, we will investigate ways of exploring more without impacting data-efficiency.

Finally, even with 12 cores, although faster than analytical approaches, Black-DROPS still requires around 25 minutes for completing 8 episodes in the cart-pole task. The main issue is the cubic computational complexity of the prediction of the GPs (we are doing around 64,000,000 GP queries per episode). Possible solutions include using local GPs~\cite{park2017patchwork,deisenroth2015distributed} or to stop using GPs and make use of recent advances in neural networks with uncertain predictions~\cite{gal2016improving,gal2015dropout}.
%
%
%
%
%
%
%

\bibliographystyle{IEEEtran}
\bibliography{IEEEabrv,mybib}

\begin{thebibliography}{10}
\providecommand{\url}[1]{#1}
\csname url@rmstyle\endcsname
\providecommand{\newblock}{\relax}
\providecommand{\bibinfo}[2]{#2}
\providecommand\BIBentrySTDinterwordspacing{\spaceskip=0pt\relax}
\providecommand\BIBentryALTinterwordstretchfactor{4}
\providecommand\BIBentryALTinterwordspacing{\spaceskip=\fontdimen2\font plus
\BIBentryALTinterwordstretchfactor\fontdimen3\font minus
  \fontdimen4\font\relax}
\providecommand\BIBforeignlanguage[2]{{%
\expandafter\ifx\csname l@#1\endcsname\relax
\typeout{** WARNING: IEEEtran.bst: No hyphenation pattern has been}%
\typeout{** loaded for the language `#1'. Using the pattern for}%
\typeout{** the default language instead.}%
\else
\language=\csname l@#1\endcsname
\fi
#2}}

\bibitem{carlson_how_2005}
J.~Carlson and R.~R. Murphy, ``How {UGVs} physically fail in the field,''
  \emph{{IEEE} Trans. on Robotics}, vol.~21, no.~3, pp. 423--437, 2005.

\bibitem{cully2015robots}
A.~Cully, J.~Clune, D.~Tarapore, and J.-B. Mouret, ``Robots that can adapt like
  animals,'' \emph{Nature}, vol. 521, no. 7553, pp. 503--507, 2015.

\bibitem{chatzilygeroudis2016resetfree}
K.~Chatzilygeroudis, V.~Vassiliades, and J.-B. Mouret, ``{Reset-free
  Trial-and-Error Learning for Data-Efficient Robot Damage Recovery},''
  \emph{arXiv:1610.04213}, 2016.

\bibitem{nagatani2013emergency}
K.~Nagatani \emph{et~al.}, ``Emergency response to the nuclear accident at the
  {Fukushima Daiichi Nuclear Power Plants} using mobile rescue robots,''
  \emph{Journal of Field Robotics}, vol.~30, no.~1, pp. 44--63, 2013.

\bibitem{mnih_human-level_2015}
V.~Mnih \emph{et~al.}, ``Human-level control through deep reinforcement
  learning,'' \emph{Nature}, vol. 518, no. 7540, pp. 529--533, 2015.

\bibitem{lecun2015deep}
Y.~LeCun, Y.~Bengio, and G.~Hinton, ``Deep learning,'' \emph{Nature}, vol. 521,
  no. 7553, pp. 436--444, 2015.

\bibitem{deisenroth_survey_2013}
M.~P. Deisenroth, G.~Neumann, and J.~Peters, ``A survey on policy search for
  robotics,'' \emph{Foundations and Trends in Robotics}, vol.~2, no.~1, pp.
  1--142, 2013.

\bibitem{kohl2004policy}
N.~Kohl and P.~Stone, ``Policy gradient reinforcement learning for fast
  quadrupedal locomotion,'' in \emph{Proc. of ICRA}, vol.~3.\hskip 1em plus
  0.5em minus 0.4em\relax IEEE, 2004, pp. 2619--2624.

\bibitem{calandra2015bayesian}
R.~Calandra, A.~Seyfarth, J.~Peters, and M.~Deisenroth, ``Bayesian optimization
  for learning gaits under uncertainty,'' \emph{Annals of Mathematics and
  Artificial Intelligence (AMAI)}, 2015.

\bibitem{lizotte2007automatic}
D.~J. Lizotte, T.~Wang, M.~H. Bowling, and D.~Schuurmans, ``Automatic gait
  optimization with {G}aussian process regression,'' in \emph{IJCAI}, vol.~7,
  2007, pp. 944--949.

\bibitem{nguyen2011model}
D.~Nguyen-Tuong and J.~Peters, ``Model learning for robot control: a survey,''
  \emph{Cognitive Processing}, vol.~12, no.~4, pp. 319--340, 2011.

\bibitem{camacho2013model}
E.~F. Camacho and C.~B. Alba, \emph{Model predictive control}.\hskip 1em plus
  0.5em minus 0.4em\relax Springer Science \& Business Media, 2013.

\bibitem{deisenroth_gaussian_2015}
M.~P. Deisenroth, D.~Fox, and C.~E. Rasmussen, ``Gaussian processes for
  data-efficient learning in robotics and control,'' \emph{IEEE Trans. Pattern
  Anal. Mach. Intell.}, vol.~37, no.~2, pp. 408--423, 2015.

\bibitem{rasmussen2006gaussian}
C.~E. Rasmussen and C.~K.~I. Williams, \emph{Gaussian processes for machine
  learning}.\hskip 1em plus 0.5em minus 0.4em\relax MIT Press, 2006.

\bibitem{hansen2001completely}
N.~Hansen and A.~Ostermeier, ``Completely derandomized self-adaptation in
  evolution strategies,'' \emph{Evolutionary computation}, vol.~9, no.~2, pp.
  159--195, 2001.

\bibitem{williams1992simple}
R.~J. Williams, ``Simple statistical gradient-following algorithms for
  connectionist reinforcement learning,'' \emph{Machine learning}, vol.~8, no.
  3-4, pp. 229--256, 1992.

\bibitem{sehnke2008policy}
F.~Sehnke \emph{et~al.}, ``Policy gradients with parameter-based exploration
  for control,'' in \emph{Proc. of Artificial Neural Networks}.\hskip 1em plus
  0.5em minus 0.4em\relax Springer, 2008, pp. 387--396.

\bibitem{kober2011power}
J.~Kober and J.~Peters, ``Policy search for motor primitives in robotics,''
  \emph{Machine Learning}, vol.~84, pp. 171--203, 2011.

\bibitem{theodorou2010generalized}
E.~Theodorou, J.~Buchli, and S.~Schaal, ``A generalized path integral control
  approach to reinforcement learning,'' \emph{JMLR}, vol.~11, pp. 3137--3181,
  2010.

\bibitem{wierstra2014natural}
D.~Wierstra \emph{et~al.}, ``Natural evolution strategies,'' \emph{JMLR},
  vol.~15, no.~1, pp. 949--980, 2014.

\bibitem{akimoto2010bidirectional}
Y.~Akimoto, Y.~Nagata, I.~Ono, and S.~Kobayashi, ``Bidirectional relation
  between {CMA} evolution strategies and natural evolution strategies,'' in
  \emph{Proc. of PPSN}.\hskip 1em plus 0.5em minus 0.4em\relax Springer, 2010,
  pp. 154--163.

\bibitem{stulp2013robot}
F.~Stulp and O.~Sigaud, ``Robot skill learning: From reinforcement learning to
  evolution strategies,'' \emph{Paladyn, Journal of Behavioral Robotics},
  vol.~4, no.~1, pp. 49--61, 2013.

\bibitem{shahriari_taking_2016}
B.~Shahriari, K.~Swersky, Z.~Wang, R.~P. Adams, and N.~de~Freitas, ``Taking the
  human out of the loop: A review of {B}ayesian optimization,'' \emph{Proc. of
  the {IEEE}}, vol. 104, no.~1, pp. 148--175, 2016.

\bibitem{auer2002using}
P.~Auer, ``Using confidence bounds for exploitation-exploration trade-offs,''
  \emph{JMLR}, vol.~3, pp. 397--422, 2002.

\bibitem{bagnell2001autonomous}
J.~A. Bagnell and J.~G. Schneider, ``Autonomous helicopter control using
  reinforcement learning policy search methods,'' in \emph{Proc. of ICRA},
  vol.~2.\hskip 1em plus 0.5em minus 0.4em\relax IEEE, 2001, pp. 1615--1620.

\bibitem{ng2006autonomous}
A.~Y. Ng, A.~Coates, M.~Diel, V.~Ganapathi, J.~Schulte, B.~Tse, E.~Berger, and
  E.~Liang, ``Autonomous inverted helicopter flight via reinforcement
  learning,'' in \emph{Experimental Robotics IX}.\hskip 1em plus 0.5em minus
  0.4em\relax Springer, 2006, pp. 363--372.

\bibitem{levine2014learning}
S.~Levine and P.~Abbeel, ``Learning neural network policies with guided policy
  search under unknown dynamics,'' in \emph{Proc. of NIPS}, 2014, pp.
  1071--1079.

\bibitem{ko2007gaussian}
J.~Ko, D.~J. Klein, D.~Fox, and D.~Haehnel, ``Gaussian processes and
  reinforcement learning for identification and control of an autonomous
  blimp,'' in \emph{Proc. of ICRA}, 2007, pp. 742--747.

\bibitem{kupcsik2014model}
A.~Kupcsik \emph{et~al.}, ``Model-based contextual policy search for
  data-efficient generalization of robot skills,'' \emph{Artificial
  Intelligence}, 2014.

\bibitem{tangkaratt2014model}
V.~Tangkaratt, S.~Mori, T.~Zhao, J.~Morimoto, and M.~Sugiyama, ``Model-based
  policy gradients with parameter-based exploration by least-squares
  conditional density estimation,'' \emph{Neural Networks}, vol.~57, pp.
  128--140, 2014.

\bibitem{ng2000pegasus}
A.~Y. Ng and M.~Jordan, ``{PEGASUS:} a policy search method for large {MDPs}
  and {POMDPs},'' in \emph{Proc. of Uncertainty in Artificial
  Intelligence}.\hskip 1em plus 0.5em minus 0.4em\relax Morgan Kaufmann, 2000,
  pp. 406--415.

\bibitem{peters2008reinforcement}
J.~Peters and S.~Schaal, ``Reinforcement learning of motor skills with policy
  gradients,'' \emph{Neural Networks}, vol.~21, no.~4, pp. 682--697, 2008.

\bibitem{cully2016limbo}
A.~Cully, K.~Chatzilygeroudis, F.~Allocati, and J.-B. Mouret, ``Limbo: A fast
  and flexible library for {B}ayesian optimization,'' \emph{arxiv:1611.07343},
  2016.

\bibitem{jin2005evolutionary}
Y.~Jin and J.~Branke, ``Evolutionary optimization in uncertain environments-a
  survey,'' \emph{IEEE Trans. on Evolutionary Computation}, vol.~9, no.~3, pp.
  303--317, 2005.

\bibitem{hansen2009method}
N.~Hansen, A.~S. Niederberger, L.~Guzzella, and P.~Koumoutsakos, ``A method for
  handling uncertainty in evolutionary optimization with an application to
  feedback control of combustion,'' \emph{IEEE Trans. on Evolutionary
  Computation}, vol.~13, no.~1, pp. 180--197, 2009.

\bibitem{miller1996genetic}
B.~L. Miller and D.~E. Goldberg, ``Genetic algorithms, selection schemes, and
  the varying effects of noise,'' \emph{Evolutionary Computation}, vol.~4,
  no.~2, pp. 113--131, 1996.

\bibitem{hansen2009benchmarking}
N.~Hansen, ``{Benchmarking a BI-population CMA-ES on the BBOB-2009 function
  testbed},'' in \emph{Proc. of GECCO}.\hskip 1em plus 0.5em minus 0.4em\relax
  ACM, 2009, pp. 2389--2396.

\bibitem{hansen2009benchmarking_noisy}
------, ``{Benchmarking a BI-population CMA-ES on the BBOB-2009 noisy
  testbed},'' in \emph{Proc. of GECCO}.\hskip 1em plus 0.5em minus 0.4em\relax
  ACM, 2009, pp. 2397--2402.

\bibitem{park2017patchwork}
C.~Park and D.~Apley, ``Patchwork kriging for large-scale gaussian process
  regression,'' \emph{arXiv preprint arXiv:1701.06655}, 2017.

\bibitem{deisenroth2015distributed}
M.~P. Deisenroth and J.~W. Ng, ``Distributed gaussian processes,''
  \emph{arXiv:1502.02843}, 2015.

\bibitem{gal2016improving}
Y.~Gal, R.~T. McAllister, and C.~E. Rasmussen, ``Improving {PILCO} with
  bayesian neural network dynamics models,'' in \emph{Data-Efficient Machine
  Learning workshop}, 2016.

\bibitem{gal2015dropout}
Y.~Gal and Z.~Ghahramani, ``Dropout as a bayesian approximation: Representing
  model uncertainty in deep learning,'' in \emph{Proc. of ICML}, 2015.

\end{thebibliography}

\end{document}